\documentclass[cameraready]{Interspeech}
\usepackage{amsmath}

\title{PRISM: Prosody-Integrated Multi-Agent Reasoning Framework for Empathetic Spoken Dialogue
}

\author[orcid=0009-0007-8962-2373]{Wen}{Zhang}
\author[orcid=0000-0001-5352-8579]{Xiaocui}{Yang}
\author[orcid=0009-0004-2435-2761]{Zhuoyue}{Gao}
\author[orcid=0000-0002-2846-7652, correspondingauthor]{Shi}{Feng}
\author[orcid=0000-0003-1340-0778]{Daling}{Wang}
\author[orcid=0000-0003-0854-2966]{Yifei}{Zhang}

\address{
   School of Computer Science and Engineering, Northeastern University, Shenyang 110819, China
}

\email{zhangw6@mails.neu.edu.cn, yangxiaocui@cse.neu.edu.cn, gaozy5@mails.neu.edu.cn, fengshi@cse.neu.edu.cn, wangdaling@cse.neu.edu.cn, zhangyifei@cse.neu.edu.cn}

\keywords{empathetic spoken dialogue, multi-agent framework, prosody-to-language translation}

\usepackage{comment}

\begin{document}

\maketitle

\begin{abstract}
    Empathetic spoken dialogue systems require not only semantically appropriate responses but also emotionally aligned prosodic expression. However, cascade pipelines often discard acoustic cues during speech-to-text conversion, while end-to-end speech models lack interpretable control over emotion and knowledge integration. To address these challenges, we propose PRISM, a multi-agent framework for empathetic spoken dialogue that decouples speech perception, response generation, and speech synthesis into coordinated components. PRISM introduces a prosody-to-language translation mechanism to stabilize large language model reasoning and enables on-demand invocation of external knowledge tools for empathetic dialogue generation. Experimental results demonstrate that PRISM achieves consistent improvements in empathy, prosodic appropriateness, and text response generation quality across objective and subjective metrics. Our code is available at: \url{https://github.com/Bxzfrm/PRISM}.

\end{abstract}

\section{Introduction}

In recent years, advances in large language models (LLMs) and speech technologies have driven human-machine dialogue systems from text-based interaction toward more natural spoken interaction. Compared with text dialogue, spoken dialogue conveys not only linguistic content but also prosodic and emotional cues \cite{eyben2015geneva, chi2025role}. These paralinguistic signals play a critical role in empathetic understanding and emotional responding \cite{zhou2018emotional, schuller2018speech}, especially in emotional support scenarios where users' affective states evolve dynamically \cite{wang-etal-2025-annaagent}. However, how to effectively perceive, model, and utilize prosodic information remains challenging. Existing studies \cite{huang2024audiogpt, serdyuk2018towards} can be categorized into cascade-based spoken dialogue systems and end-to-end speech models.

Traditional spoken dialogue systems typically adopt an “ASR-text dialogue model-TTS” cascade paradigm \cite{ji2024wavchat}. While relatively mature, once speech is transcribed into text, prosodic information related to emotion and expressiveness is inevitably lost. Although the system may generate semantically appropriate responses, it often fails to exhibit genuine empathy at the speech level \cite{eyben2010opensmile}. Moreover, the lack of consistency constraints between emotional intent and speech synthesis leaves empathy largely confined to the textual domain. Recently, end-to-end speech models have emerged, aiming to directly model mappings from speech to speech or from speech to multimodal outputs \cite{borsos2022audiolm, speechgpt, chen2025valle, Rubenstein2023AudioPaLMAL}. While these approaches exhibit strong representational capacity and high naturalness in generation, they typically treat prosody and emotion as implicit features, lacking interpretable intermediate representations. In addition, such models often rely on large-scale joint training, making integration of new knowledge or emotional strategies inflexible.

Knowledge augmentation improves the factuality and contextual adaptability of dialogue systems \cite{lewis2020rag}, and LLMs can dynamically acquire external information through tool calling without additional training \cite{Yao2022ReActSR}. However, prior work focuses on text-based settings. 
Empathetic spoken dialogue requires jointly modeling prosodic perception, emotional reasoning, knowledge augmentation, and speech generation across modalities. Conventional pipelines often suffer from irreversible information loss and accumulated errors due to strictly sequential processing. In contrast, a tool-calling multi-agent paradigm enables feedback-driven coordination and adaptive tool use through structured intermediate representations, allowing emotional intent and contextual knowledge to propagate from perception to generation.

To address these challenges, we propose PRISM (Prosody-aware Reasoning Integrated System via a Multi-agent framework), which decouples speech perception, dialogue management, response generation, and speech synthesis into specialized agents. To facilitate stable emotional reasoning over prosodic signals, the framework incorporates a prosody-to-language translation mechanism converting acoustic cues into interpretable textual representations. PRISM supports on-demand invocation of external knowledge tools during dialogue, allowing plug-and-play updates of knowledge sources by modifying the external tools without retraining the underlying models. In summary, our contributions are as follows:

\begin{itemize}
  \item We propose PRISM, a multi-agent framework with feedback-driven coordination that enables prosody-aware dialogue reasoning and flexible knowledge integration for empathetic spoken dialogue.
  \item We introduce a prosody-to-language translation mechanism that maps acoustic and temporal cues into interpretable natural-language descriptions, enabling LLM to reason about user’s states in a more stable and human-aligned manner.
  \item We conduct extensive experiments on the AvaMERG dataset and observe that PRISM consistently outperforms baseline models across automatic, human, and LLM-based evaluations. Ablation studies further validate the effectiveness of each proposed component.

\end{itemize}

\begin{figure*}[t]
  \centering
  \vspace{-8pt}
  \includegraphics[width=0.92\linewidth]{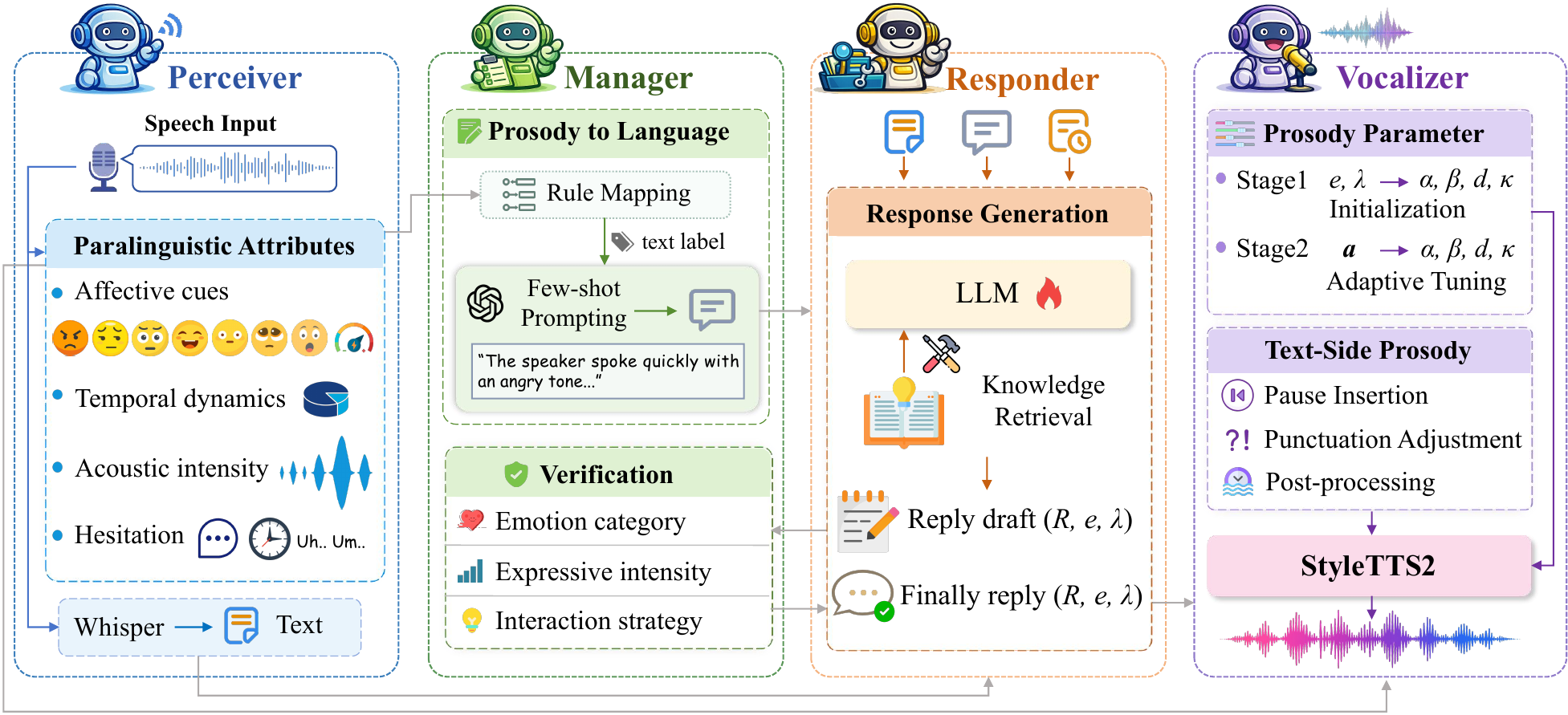}
  \vspace{-5pt}
  \caption{Overview of the PRISM Framework.}
  \label{fig:framework}
\end{figure*}

\section{Method}
We propose a prosody-aware multi-agent empathetic spoken dialogue framework consisting of four components: Perceiver, Manager, Responder, and Vocalizer, as illustrated in Figure~\ref{fig:framework}.

\subsection{Perceiver}
For a given input speech $x$, Perceiver outputs a structured state $s = \{T, \mathbf{a}\}$, where $T$ is the transcription text, $\mathbf{a}$ represents a set of paralinguistic attributes designed to capture the speaker’s emotional and expressive state. The paralinguistic attributes 
$\mathbf{a}$ include four categories of cues: (i) affective cues, including the recognized emotion category and its confidence score; (ii) temporal dynamics, including speaking rate and pause ratio; (iii) acoustic intensity, characterized by energy statistics; (iv) disfluency-related cues, including filler rate and an interpretable expression certainty score. 

We employ OpenAI Whisper \cite{radford2023whisper} to convert the raw speech waveform into text. Utterance-level emotion is classified using FunASR's emotion2vec \cite{ma-etal-2024-emotion2vec}, a self-supervised speech emotion recognizer fine-tuned on multi-domain emotional corpora. The model outputs:
\begin{itemize}
  \item Emotion label $y \in \{\text{angry},\allowbreak
\text{disgusted},\allowbreak
\text{fearful},\allowbreak
\text{happy},\allowbreak
\text{neutral},\allowbreak
\text{other},\allowbreak
\text{sad},\allowbreak
\text{surprised},\allowbreak
\text{melodious},\allowbreak
\text{unknown}\}$.
  \item Confidence score $q_y$: softmax probability of the top-1 class.
\end{itemize}
To capture temporal speaking behavior, we resample audio to 16~kHz and estimate speech and silence durations using WebRTC voice activity detector (VAD). The pause ratio $p$ is computed as the proportion of silence duration to the total utterance duration. The speaking rate is defined as the ratio between the number of transcript tokens and the estimated speech duration,
\begin{equation}
r = \frac{N_{\text{tok}}(T)}{t_{\text{sp}}}, 
\end{equation}
where $N_{\text{tok}}(T)$ denotes the number of tokens in the transcription and $t_{\text{sp}}$ denotes the speech duration.
To characterize acoustic intensity, we compute frame-level RMS energy over the utterance and summarize it using the mean and standard deviation, denoted as $(\mu_E, \sigma_E)$. These statistics provide a compact representation of the speaker’s overall loudness and energy variation.
Disfluency and hesitation are quantified using the filler rate, defined as:
\begin{equation}
f = \frac{N_f}{\max\left(N_{\text{tok}}(T),\, 1\right)}, 
\end{equation}
where $N_f$ denotes the number of predefined filler expressions detected in the transcription. In addition, to quantify speaker confidence, we derive a heuristic certainty score
$c \in [0,1]$ based on speaking rate, pause behavior, and disfluency. Specifically, we first normalize
the speaking rate $r$, pause ratio $p$, and filler rate $f$ as:
\begin{align}
\label{eq:normalized_prosody}
\tilde{r} &= \mathrm{clip}\left(\frac{r - 2}{4}, 0, 1\right),\\
\tilde{p} &= \mathrm{clip} (p, 0, 1),\\
\tilde{f} &= \mathrm{clip}\left(\frac{f}{0.2}, 0, 1\right).
\end{align}
where $\mathrm{clip}(x,0,1)$ bounds $x$ to the interval $[0,1]$. The certainty score is then computed as a weighted combination:
\begin{equation}
c = \mathrm{clip}\left(0.55\,\tilde{r} + 0.25\,(1 - \tilde{p}) + 0.20\,(1 - \tilde{f}),\, 0,\, 1\right).
\end{equation}

Perceiver converts raw speech into a structured paralinguistic state that can be used for downstream dialogue and empathetic response generation.

\subsection{Manager}
Manager acts as a central coordination module responsible for controlling the dialogue flow. Its role is reflected in the following two aspects.

\textbf{Prosody-to-language translation}. Given the structured paralinguistic attributes produced by Perceiver, Manager converts numerical and categorical prosodic cues into a concise natural-language description $D$. The process is divided into two stages. First, numerical attributes are mapped to descriptive textual labels via rule-based thresholding, yielding a stable and interpretable intermediate representation. Second, an LLM is employed under a few-shot prompting strategy to generate a coherent and natural prosody description. The prompt template \cite{wang2024langgptrethinkingstructuredreusable} includes carefully designed examples that illustrate how factors such as emotional intensity, rhythm, pausing behavior, speech energy, and certainty jointly contribute to natural prosodic expression. By converting paralinguistic information into language-level descriptions, Manager enables Responder to perceive the user’s expressive state in a manner closer to human understanding, without directly processing low-level acoustic numerical features.

\textbf{Response-level verification}. To enhance generation stability and expressive consistency, we introduce a lightweight alignment-check module within Manager to perform post-hoc verification of Responder’s outputs. This module evaluates whether a response aligns with the user’s prosodic description in terms of emotion category, expressive intensity, and interaction strategy. When inconsistencies are detected, it provides minimal revision suggestions to guide Responder’s refinement.

\subsection{Responder}
Responder is responsible for generating the final empathetic response at the dialogue level and incorporating external knowledge when necessary. Unlike cascade approaches that model tool invocation and text generation as separate stages, we adopt a unified language model that jointly handles knowledge usage decisions and empathetic response generation, thereby reducing error propagation across modules.

Specifically, Responder takes the current transcript $T$, the language prosody description $D$ generated by Manager, and the dialogue history $H$ as inputs. LLM first implicitly determines whether external knowledge support is required; when it detects that commonsense or contextual enrichment would lead to a more appropriate response, the system retrieves multi-dimensional commonsense information via external tools and injects it into the generation context in textual form. This design enables knowledge expansion by replacing or updating the knowledge interfaces, without retraining model parameters.

During response generation, the model uses the prosody description as an explicit conditioning signal to dynamically adjust its response strategy and emotional expression. For example, when the prosodic description indicates user hesitation or low emotional valence, the model tends to generate more reassuring and supportive responses; in contrast, in high-arousal emotional scenarios, it correspondingly amplifies emotional resonance. In addition to the response text $R$, Responder also predicts the target emotion category $e$ and its expressive intensity $\lambda$, providing control signals for subsequent emotionally adaptive speech synthesis.

\subsection{Vocalizer}
Vocalizer converts textual responses into expressive speech that mirrors the emotional context through adaptive prosodic control. We employ StyleTTS2 \cite{styletts2}, a diffusion-based TTS model supporting reference-based voice cloning and fine-grained prosody manipulation.

\textbf{Prosody Parameter Computation}. Vocalizer computes speech synthesis parameters via a two-stage control process, integrating both target empathetic intent and user’s prosodic cues. Four parameters are dynamically determined: timbre similarity $\alpha$, prosody strength $\beta$, the number of diffusion refinement steps $d$, and an expressive scaling factor $\kappa$.
\begin{equation}
(\alpha, \beta, d, \kappa) = V(e, \lambda, \mathbf{a}).
\end{equation}
In the first stage, base values are initialized according to the target emotion $e$ and its expressive intensity $\lambda \in [0,1]$, as predicted by Responder. The emotion intensity $\lambda$ primarily influences prosody strength and synthesis fidelity by adjusting $\beta$ and $d$: higher intensity leads to stronger prosodic variation and more refinement steps, while lower intensity yields flatter and more neutral delivery. The emotion category $e$ controls the expressive scaling factor $\kappa$, enabling different affective styles. In the second stage, the base parameters are adaptively refined using the user paralinguistic attributes $\mathbf{a}$ output by Perceiver. Specifically, features such as expressive certainty, emotional intensity, and energy-related statistics are used to modulate prosody realization in an empathy-aware manner. When the user exhibits low certainty or frequent pauses, the synthesizer reduces prosody strength $\beta$ and attenuates the expressive scaling factor $\kappa$ to produce a gentler delivery; when strong negative emotions are detected, $\beta$ is increased to better match the required level of empathetic expressiveness. Finally, all parameters are clipped to valid ranges to ensure synthesis stability.

\textbf{Text-Side Prosody Shaping and Post-Processing}. For users with frequent pauses, short pause markers are inserted after punctuation to yield a more patient speaking rhythm. For highly positive target emotions, punctuation is adjusted to enhance expressiveness, whereas for sad or soothing responses, excessive exclamation is suppressed to maintain a calm tone. Finally, optional post-processing is applied to further adjust speaking rate and output energy through time-stretching and amplitude scaling when required. These operations allow fine-grained alignment between the synthesized speech and the user’s conversational rhythm.

\section{Experiments}

\begin{table*}[t]
\centering
\caption{Comparison results between PRISM and other baseline models on the AvaMERG test set}
\small
\setlength{\tabcolsep}{8pt}
\renewcommand{\arraystretch}{0.95}  
\begin{tabular}{lcccc}
\toprule
 \textbf{Model} & \textbf{ROU-1/2/L} & \textbf{B-S} & \textbf{BLEU-1/2/3/4} & \textbf{Dist-1/2} \\
\midrule
ASR+LLM          & 0.1690/0.0271/0.1406 & 0.8652 & 0.1431/0.0514/0.0250/0.0132& 0.0286/0.1722 \\
SpeechGPT        & 0.1437/0.0228/0.1126 & 0.8534 & 0.1189/0.0543/0.0304/0.0180 & 0.0306/0.1340 \\
OSUM-EChat       & 0.1546/0.0263/0.1146 & 0.8673 & 0.1381/0.0485/0.0226/0.0111 & 0.0342/0.2133 \\
SALMONN-7B       & 0.1598/0.0321/0.1226 & 0.8684 & 0.1415/0.0616/0.0357/0.0217 & 0.0225/0.1346 \\
SALMONN-13B      & 0.1666/0.0381/0.1289 & 0.8705 & 0.1464/0.0570/0.0303/0.0174 & 0.0218/0.1327 \\
Qwen2.5-Omni-7B  & 0.1880/0.0542/0.1555 & 0.8746 & 0.1737/0.0831/0.0530/0.0352 & 0.0375/0.2380 \\
LLaMA-Omni2     & 0.1703/0.0329/0.1330 & 0.8674 & 0.1565/0.0646/0.0354/0.0205 & \textbf{0.0460}/0.2376 \\ 
OpenS2S          & 0.1759/0.0356/0.1408 & 0.8691 & 0.1883/0.0700/0.0355/0.0192 & 0.0428/0.2329 \\

\midrule
\textbf{PRISM (Qwen)}  & \textbf{0.2254/0.0745/0.1872}  & \underline{0.8792} & \underline{0.2041}/\underline{0.1142}/\underline{0.0792}/\textbf{0.0571} & 0.0390/\underline{0.2519} \\
\textbf{PRISM (Llama)} & 0.2027/0.0649/0.1743 & \textbf{0.8801} & \textbf{0.2318/0.1223/0.0805}/0.0555 & 0.0409/\textbf{0.2574} \\
\midrule
Always Kno  & 0.1611/0.0204/0.1301 & 0.8667 & 0.1606/0.0560/0.0259/0.0133 & 0.0258/0.1550 \\
w/o Kno     & 0.1624/0.0205/0.1300 & 0.8633 & 0.1601/0.0549/0.0254/0.0129 & 0.0250/0.1495  \\
w/o Prosody-Desc  & 0.1521/0.0202/0.1262 & 0.8641 & 0.1727/0.0583/0.0271/0.0137 &  0.0280/0.1617  \\

\bottomrule
\end{tabular}
\label{tab:main_results}
\end{table*}

\subsection{Dataset}
\textbf{TOOL-ED} \cite{tooled} is a tool-augmented extension of the ED \cite{ed} dataset, designed to study empathetic dialogue generation with external knowledge integration. Each sample consists of a short dialogue context and its corresponding empathetic response, along with annotations indicating whether external knowledge should be invoked. \textbf{AvaMERG} \cite{avamerg} is a multimodal empathetic dialogue dataset that extends ED with speech and facial expression annotations. We evaluate our system on the standard audio subset split of AvaMERG.

\subsection{Baselines}
To validate the effectiveness of our PRISM, we select the following baseline models for comparison and conduct a comparative evaluation based on test results on the AvaMERG dataset.

\textbf{ASR+LLM}: A cascaded pipeline using OpenAI Whisper for transcription followed by an LLM for response generation. 
\textbf{SpeechGPT} \cite{speechgpt}: An end-to-end speech dialogue system integrating speech understanding and generation. 
\textbf{SALMONN} \cite{tang2024salmonn}: A dual-encoder model for joint speech understanding and response generation. We report results for both 7B and 13B. 
\textbf{OSUM-EChat} \cite{geng2025osumechat}: An emotion-aware speech dialogue system generating responses from speech inputs. 
\textbf{Qwen2.5-Omni-7B} \cite{xu2025qwen25omni}: A multimodal model supporting unified speech, vision, and text dialogue.
\textbf{LLaMA-Omni2} \cite{llamaomni}: A speech-enabled LLaMA-style model for end-to-end multimodal dialogue.
\textbf{OpenS2S} \cite{wang-etal-2025-opens2s}:  An open-source speech-to-speech dialogue model mapping input speech directly to output speech. 

\subsection{Implementation Details} 
We fine-tune Qwen2.5-7B-Instruct \cite{qwen2025qwen25technicalreport} and Llama-3.1-8B-Instruct \cite{grattafiori2024llama3herdmodels} on TOOL-ED to serve as Responder. Among all modules in PRISM, only the Responder is fine-tuned. Training is conducted using the LLaMA-Factory \cite{zheng-etal-2024-llamafactory} on NVIDIA A6000 (48GB) GPUs. We use GPT-3.5-Turbo via the OpenAI API as Manager. For external knowledge augmentation, we employ COMET-BART \cite{bosselut-etal-2019-comet}, a BART-based commonsense generation model. All baseline models are fine-tuned on the AvaMERG training set using the hyperparameter settings recommended in their original papers or official implementations.

\subsection{Results and Analysis}

\subsubsection{Main Results}
We present comparative results through automatic evaluation, LLM-based evaluation, and human evaluation.

\textbf{Automatic Evaluation}. Table~\ref{tab:main_results} presents the automatic evaluation results on the AvaMERG test set. The evaluation is conducted across four categories of metrics: n-gram overlap (BLEU) \cite{papineni-etal-2002-bleu}, semantic similarity (BERTScore) \cite{zhang2019bertscore}, content overlap (ROUGE) \cite{lin-2004-rouge}, and lexical diversity (Dist) \cite{li-etal-2016-diversity}. Overall, PRISM (Qwen) and PRISM (Llama) outperform all baseline models across nearly all evaluation metrics, particularly achieving substantial improvements in ROUGE and BLEU scores, demonstrating PRISM's advantage in generation quality and content relevance.

\begin{figure}[!h]
    \centering
    \vspace{-5pt}
    \includegraphics[width=0.8\linewidth]{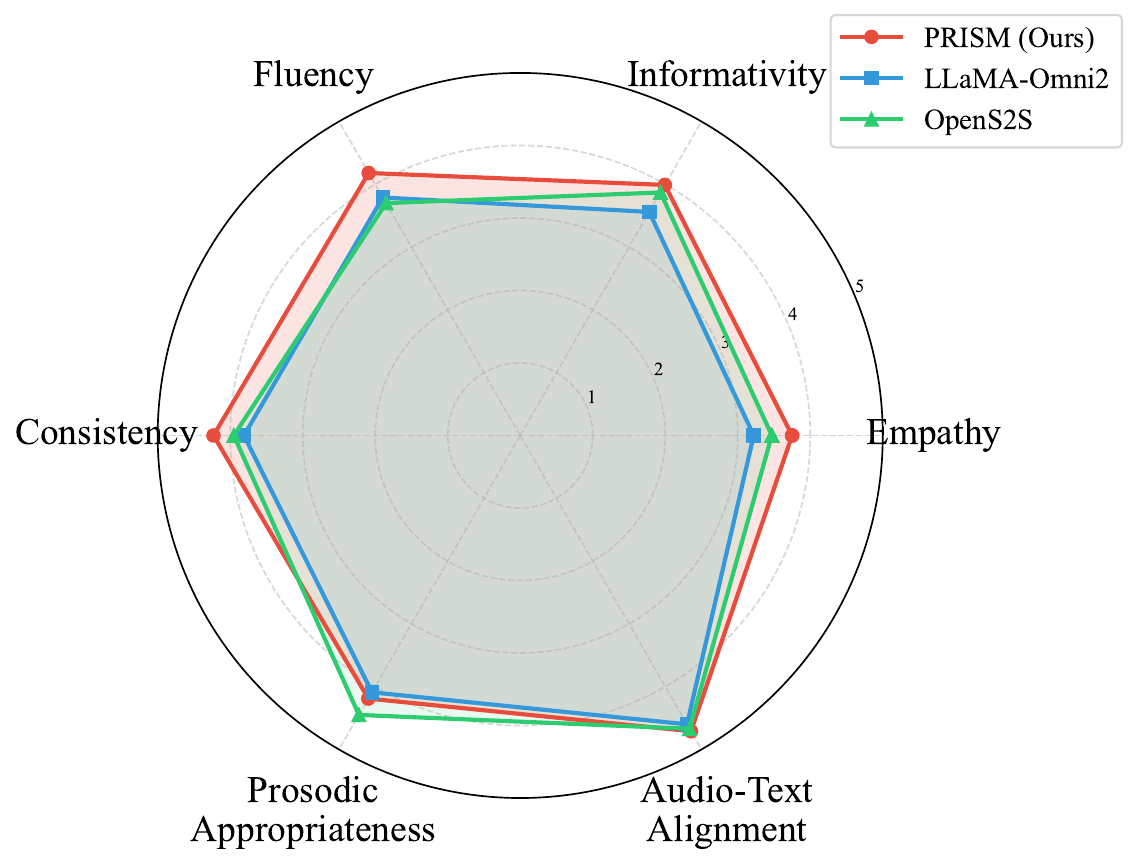}
    \vspace{-5pt}
    \caption{Human evaluation results.}
    \label{fig:human_evaluation}
\end{figure}

\textbf{Human Evaluation}. We recruited three researchers specializing in empathetic dialogue systems as annotators. A total of 100 dialogue samples were randomly selected for evaluation. Each sample was independently rated by all annotators. The evaluation is conducted from two aspects, text quality and speech quality, covering the following six dimensions: Empathy, Informativity, Fluency, Consistency, Prosodic Appropriateness, Audio-Text Alignment. All evaluations are conducted using a 5-point Likert scale \cite{likert1932technique}. Inter-annotator agreement was measured using intraclass correlation coefficient (ICC), yielding an average score of 0.81, indicating substantial agreement. As shown in Figure~\ref{fig:human_evaluation}, PRISM demonstrates comparable or superior performance across most evaluation dimensions when compared with the advanced speech models, LLaMA-Omni2 and OpenS2S.

\begin{figure}[!h]
    \centering
    \vspace{-5pt}
    \includegraphics[width=0.85\linewidth]{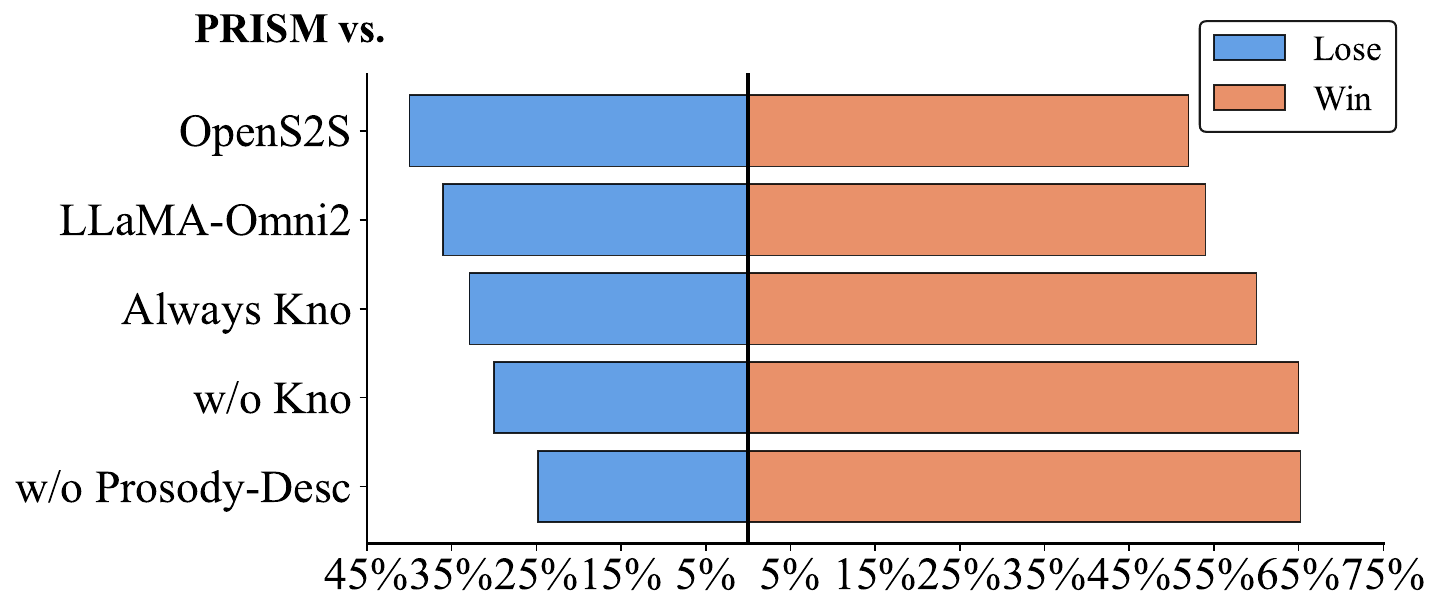}
    \vspace{-5pt}
    \caption{LLM-based evaluation results. Only wins and losses are shown; ties are omitted.}
    \label{fig:LLM-based_evaluation}
\end{figure}

\textbf{LLM-based Evaluation}. To comprehensively evaluate the quality of the generated responses, we additionally employ GPT-4o for automatic evaluation. The model conducts A/B testing \cite{sabour2022cem} primarily from three aspects, empathy, fluency, and consistency to compare the performance of PRISM against other baselines. As shown in Figure~\ref{fig:LLM-based_evaluation}, PRISM demonstrates a higher win rate against OpenS2S and LLaMA-Omni2.

\subsubsection{Ablation Studies}
We conducted ablation studies on two key aspects: autonomous knowledge invocation and prosody description. Specifically, we evaluated the results under the following conditions: enforcing knowledge usage at every turn (Always Kno), excluding knowledge entirely (w/o Kno) and removing prosody descriptions (w/o Prosody-Desc). All ablation experiments were performed based on Qwen2.5-7B-Instruct. As demonstrated 
in Table~\ref{tab:main_results} and Figure~\ref{fig:LLM-based_evaluation}, all ablation variants exhibit performance degradation, thereby validating the effectiveness of our proposed method.

\section{Conclusion}
We present PRISM, a multi-agent framework for empathetic spoken dialogue that integrates prosody-to-language translation and adaptive knowledge invocation. By decoupling perception, reasoning, and synthesis, PRISM enables interpretable emotion modeling and controllable speech generation, achieving improved empathy and prosodic alignment over baselines.

\section{Acknowledgments}
The work was supported by the National Natural Science Foundation of China (Nos. 62272092, 62172086), and the Fundamental Research Funds for
the Central Universities under Grants (N25XQD004). Thanks to the KinaMind society for their inspiring environment and unwavering support.




\section{Generative AI Use Disclosure}
Generative AI tools were used only for proofreading and language polishing. All technical contributions, experimental design, and analyses were conducted by the authors.

\bibliographystyle{IEEEtran}
\bibliography{mybib}

\end{document}